\documentclass{article}
\usepackage[preprint]{neurips_2026}

\usepackage{amsmath,amssymb,amsthm}
\usepackage{graphicx}
\usepackage{url}
\usepackage{hyperref}
\usepackage{listings}
\usepackage{xcolor}
\usepackage{booktabs}
\usepackage{algorithm}
\usepackage{algpseudocode}
\usepackage{tikz}
\usetikzlibrary{shapes,arrows,positioning,fit,backgrounds,calc,patterns,decorations.markings}
\usepackage{pgfplots}
\pgfplotsset{compat=1.18}
\usepgfplotslibrary{groupplots}
\usepackage{subcaption}
\usepackage{pifont}
\usepackage{multirow}

\newtheorem{definition}{Definition}

\title{Adaptive Vision-Language Model Routing for Computer Use Agents}

\author{
  Xunzhuo Liu$^{1}$, Bowei He$^{2,3,}$\thanks{Corresponding author: \texttt{Bowei.He@mbzuai.ac.ae}} , Xue Liu$^{2,3}$, Andy Luo$^{4}$, Haichen Zhang$^{4}$, Huamin Chen$^{5}$ \\
  $^1$ vLLM Semantic Router Project, $^2$ MBZUAI, $^3$ McGill University, $^4$ AMD, $^5$ Red Hat
}

\begin{document}
\maketitle

\begin{abstract}

Computer Use Agents (CUAs) translate natural-language instructions into Graphical User Interface (GUI) actions such as clicks, keystrokes, and scrolls by relying on a Vision-Language Model (VLM) to interpret screenshots and predict grounded tool calls. However, grounding accuracy varies dramatically across VLMs, while current CUA systems typically route every action to a single fixed model regardless of difficulty. We propose \textbf{Adaptive VLM Routing} (AVR), a framework that inserts a lightweight semantic routing layer between the CUA orchestrator and a pool of VLMs. For each tool call, AVR estimates action difficulty from multimodal embeddings, probes a small VLM to measure confidence, and routes the action to the cheapest model whose predicted accuracy satisfies a target reliability threshold. For \textit{warm} agents with memory of prior UI interactions, retrieved context further narrows the capability gap between small and large models, allowing many actions to be handled without escalation. We formalize routing as a cost--accuracy trade-off, derive a threshold-based policy for model selection, and evaluate AVR using ScreenSpot-Pro grounding data together with the OpenClaw agent routing benchmark. Across these settings, AVR projects inference cost reductions of up to 78\% while staying within 2 percentage points of an all-large-model baseline. When combined with the Visual Confused Deputy guardrail, AVR also escalates high-risk actions directly to the strongest available model, unifying efficiency and safety within a single routing framework. Materials are also provided\footnote{Model, benchmark, and code: \url{https://github.com/vllm-project/semantic-router}}.

\end{abstract}

\section{Introduction}\label{sec:intro}
\begin{figure}[t]
\centering
\begin{tikzpicture}[
  node distance=1.0cm,
  stage/.style={rectangle, draw, rounded corners=4pt, minimum width=2.0cm,
                minimum height=0.7cm, align=center, font=\small\bfseries},
  cost/.style={font=\scriptsize\itshape, text=red!70!black},
  arrow/.style={->, thick, >=stealth},
  looparrow/.style={->, thick, >=stealth, dashed, gray}
]

\node[stage, fill=blue!15] (observe) {Observe};
\node[stage, fill=orange!15, right=1.2cm of observe] (reason) {Reason};
\node[stage, fill=green!12, right=1.2cm of reason] (act) {Act};

\draw[arrow] (observe) -- node[above, font=\scriptsize] {screenshot} (reason);
\draw[arrow] (reason) -- node[above, font=\scriptsize] {tool call} (act);
\draw[looparrow, bend left=35] (act) to node[below, font=\footnotesize] {next step} (observe);

\node[cost, below=0.3cm of observe] {$\sim$2--5K tokens};
\node[cost, below=0.3cm of reason] {VLM inference};
\node[cost, below=0.3cm of act] {$(x,y)$ + action};

\node[draw, dashed, rounded corners, inner sep=8pt, fit=(observe)(reason)(act),
      label={[font=\scriptsize\bfseries]above:Repeated 5--20$\times$ per task}] {};

\node[right=0.6cm of act, font=\scriptsize, align=left, text=red!70!black]
  (costnote) {Total per task:\\400K input tokens\\$\sim$\$0.16--\$0.40};
\end{tikzpicture}
\caption{The CUA action loop.  Each iteration requires a full VLM
inference call carrying screenshots (2--5K tokens each) plus context.
A 20-step task accumulates ${\sim}$400K input tokens.}
\label{fig:cua_loop}
\end{figure}
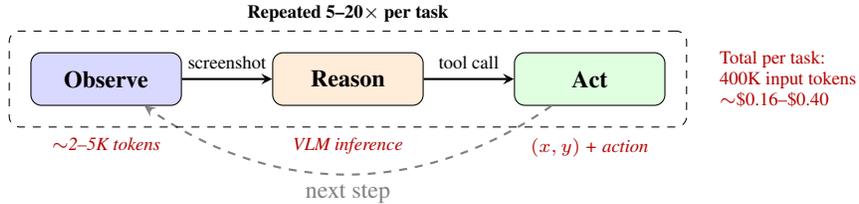

The emergence of Computer Use Agents (CUAs), systems that control a desktop Graphical User Interface (GUI) on behalf of a user by interpreting screenshots and emitting tool calls—has transformed vision-language models (VLMs) from passive question-answerers into active \emph{actuators} in the digital world. Recent systems such as OpenAI's Computer-Using Agent~\citep{openai_cua2025}, Anthropic's Computer Use tool~\citep{anthropic_cu2024}, and open-source platforms like UFO2~\citep{ufo2_2025} demonstrate that modern VLMs can perform multi-step computer tasks by repeatedly observing the screen, reasoning about the task state, and executing actions. In these systems, the agent operates in a simple iterative loop: it captures the current screen, sends the screenshot and task context to a VLM, interprets the
model’s response as a tool call (e.g., a click at $(x,y)$, a keystroke, a scroll), executes the action, and repeats until the task is complete, as illustrated in Figure~\ref{fig:cua_loop}.

Despite their impressive capabilities, current CUA systems typically rely on a single frontier VLM for every step of the action loop. This design introduces significant cost and latency overhead because each iteration carries a large multimodal context window containing screenshots, prompts, and action history. A multi-step task may require 5--20 iterations, resulting in hundreds of thousands of input tokens per task. In practice, running all actions through a large VLM such as GPT-4o~\citep{gpt4o2024}, Claude~3.5 Sonnet~\citep{anthropic_claude35sonnet_2024}, or Qwen2.5-VL-72B~\citep{qwen25vl2025} can cost on the order of \$0.10--\$0.40 per task, making large-scale deployment of CUAs expensive.

At the same time, increasing model size does not reliably improve GUI grounding performance. ScreenSpot-Pro~\citep{screenspotpro2025} shows that grounding accuracy varies widely across models and does not scale smoothly with parameter count: GPT-4o achieves only 0.8\% accuracy, while the 7B specialist OS-Atlas reaches 18.9\%. Even the strongest model evaluated, Qwen2.5-VL-72B, achieves only
43.6\%~\citep{qwen25vl2025}. Furthermore, within the same model family, scaling from 7B to 72B yields diminishing returns relative to the increase in parameters. A key observation is that CUA actions vary substantially in difficulty: clicking a large “Submit” button is trivial, while locating a small icon in a dense IDE toolbar can be challenging. Consequently, using a single model for all actions is inefficient—large models are unnecessarily expensive for easy actions, yet smaller models may fail on harder ones.

These observations suggest that CUA inference should be treated as a \emph{routing problem}. Instead of relying on a fixed model for all actions, an agent could dynamically select the cheapest model that is likely to succeed on the current step.  Such an approach would allow easy actions to be handled by lightweight models while reserving frontier VLMs for the most challenging cases. To this end, we propose \textbf{Adaptive VLM Routing (AVR)}, a framework that selects the most cost-effective model for each CUA tool call. AVR introduces three key mechanisms:
\begin{enumerate}
\item \textbf{Difficulty classification} via multimodal embeddings
      (Section~\ref{sec:difficulty}).  A lightweight 120M-parameter model
      estimates the difficulty of each action using the screenshot
      region and the action description.
\item \textbf{Confidence-based routing} via logprob probing
      (Section~\ref{sec:confidence}).  A small VLM is first queried and
      its log-probability confidence is used to determine whether its
      prediction should be accepted or escalated to a larger model.
\item \textbf{Memory-compensated routing} for warm agents
      (Section~\ref{sec:memory}).  Retrieved interaction history
      provides additional UI context, allowing the small VLM to handle
      actions that would otherwise require escalation.
\end{enumerate}

Furthermore, when integrated with the Visual Confused Deputy guardrail \citep{visualconfuseddeputy2026}, AVR can incorporate safety signals into the same routing decision.  Actions flagged as potentially dangerous can be automatically escalated to a stronger model for additional verification, enabling a unified treatment of cost optimization and safety (Section~\ref{sec:safety_routing}). Overall, AVR reframes computer-use inference as a dynamic allocation of model capacity across actions.  By routing each step to the cheapest model capable of solving it, the framework aims to significantly reduce inference cost while preserving the grounding performance required for reliable computer-use agents.

\section{Problem Formulation}\label{sec:formulation}

\subsection{CUA Tool-Call Model}
A CUA session consists of a sequence of tool calls
$\mathcal{T} = (t_1, t_2, \ldots, t_N)$, where each tool call $t_i$ is a tuple:
\begin{equation}
  t_i = (\mathbf{s}_i, \mathbf{h}_i, a_i, \mathbf{p}_i),
\end{equation}
where $\mathbf{s}_i \in \mathbb{R}^{H \times W \times 3}$ is the screenshot, $\mathbf{h}_i$ is the action history, $a_i$ is the action type (click, type, scroll), and $\mathbf{p}_i = (x_i, y_i)$ are the predicted pixel coordinates.

Given a pool of $K$ VLMs $\mathcal{M} = \{m_1, \ldots, m_K\}$ ordered by increasing capability (and cost), the routing problem is to select model $m_k$ for each tool call $t_i$ such that the total cost is minimized subject to an accuracy constraint:

\begin{equation}\label{eq:routing_objective}
\min_{\pi} \sum_{i=1}^{N} c_{\pi(i)} \quad
\text{s.t.} \quad
\frac{1}{N} \sum_{i=1}^{N}
\mathbb{I}[\text{correct}(t_i, m_{\pi(i)})] \geq \tau_{\text{acc}},
\end{equation}
where $\pi: \{1,\ldots,N\} \to \{1,\ldots,K\}$ is the routing policy, $c_k$ is the per-token cost of model $m_k$, and $\tau_{\text{acc}}$ is the target accuracy.

\subsection{Difficulty as a Latent Variable}
We hypothesize that each tool call has a latent difficulty $d_i \in [0, 1]$ that determines the minimum model capability required for correct grounding.  Formally, for each model $m_k$ there exists a
capability threshold $\theta_k$ such that:
\begin{equation}
  P[\text{correct}(t_i, m_k)] \approx \sigma\!\left(\frac{\theta_k - d_i}{\gamma}\right),
\end{equation}
where $\sigma$ is the sigmoid function and $\gamma$ controls the sharpness of the transition.  The routing problem reduces to estimating $d_i$ and selecting the cheapest model whose capability exceeds it:
\begin{equation}\label{eq:optimal_policy}
  \pi^*(i) = \min\{k : \theta_k \geq d_i + \Delta\}
\end{equation}
where $\Delta$ is a safety margin.

\subsection{Cost Model}
For a pool of two models (small $S$ and large $L$), the expected cost per tool call under a routing policy with escalation rate $\alpha$ is:
\begin{equation}\label{eq:cost_model}
  \mathbb{E}[c] = (1 - \alpha)\,c_S + \alpha\,(c_S^{\text{probe}} + c_L),
\end{equation}
where $c_S^{\text{probe}}$ is the cost of the non-streaming probe to the small model (typically ${\sim}$10\% of a full generation due to shorter output), and $c_L$ is the cost of a full generation on the large model.  The cost saving relative to all-large is:
\begin{equation}
  \text{savings} = 1 - \frac{\mathbb{E}[c]}{c_L}
  = (1 - \alpha)\left(1 - \frac{c_S}{c_L}\right) - \alpha\,\frac{c_S^{\text{probe}}}{c_L}.
\end{equation}
When the small model is $10\times$ cheaper ($c_S/c_L = 0.1$) and probing adds 10\% overhead, achieving $\alpha = 0.2$ (escalating only 20\% of actions) yields 70\% cost reduction.

\section{Adaptive VLM Routing}\label{sec:method}
\begin{figure}[t]
\centering
\begin{tikzpicture}[
  node distance=0.8cm and 1.2cm,
  box/.style={rectangle, draw, rounded corners, minimum width=2.2cm,
              minimum height=0.8cm, align=center, font=\small},
  arrow/.style={->, thick, >=stealth},
  every node/.style={font=\small}
]
\node[box, fill=blue!10] (cua) {CUA\\Orchestrator};
\node[box, fill=orange!15, right=1.5cm of cua] (router) {Semantic\\Router};
\node[box, fill=green!10, above right=0.6cm and 1.5cm of router] (small) {Small VLM\\(7B)};
\node[box, fill=red!10, below right=0.6cm and 1.5cm of router] (large) {Large VLM\\(72B)};
\node[box, fill=purple!10, below=0.8cm of router] (memory) {Memory\\Store};
\node[box, fill=yellow!10, above=0.8cm of router] (embed) {Multimodal\\Embedder};
\node[box, fill=gray!10, left=0.8cm of memory] (kb) {Contrastive\\KB};

\draw[arrow] (cua) -- node[above] {tool call} (router);
\draw[arrow] (router) -- node[above, sloped] {conf $\geq \tau$} (small);
\draw[arrow] (router) -- node[below, sloped] {conf $< \tau$} (large);
\draw[arrow, dashed] (router) -- (memory);
\draw[arrow, dashed] (router) -- (embed);
\draw[arrow, dashed] (kb) -- (router);
\draw[arrow, dotted, bend left=30] (small) to node[right, xshift=0.2cm] {logprobs} (router);
\end{tikzpicture}
\caption{AVR architecture.  The semantic router intercepts CUA tool
calls, classifies difficulty via the multimodal embedder, probes the
small VLM for confidence, injects memories for warm agents, and routes
to the cheapest sufficient model.  The contrastive KB (from the Visual
Confused Deputy guardrail) can force escalation for high-risk actions.}
\label{fig:architecture}
\end{figure}
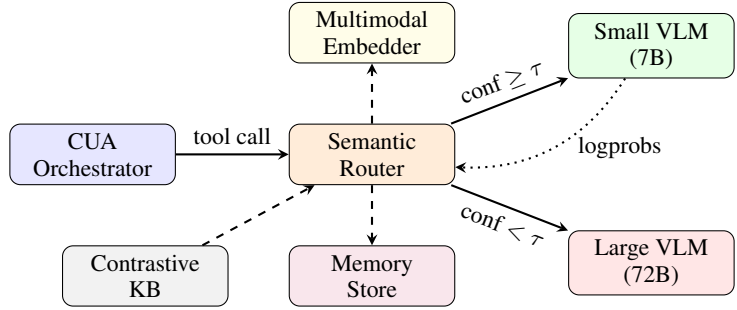

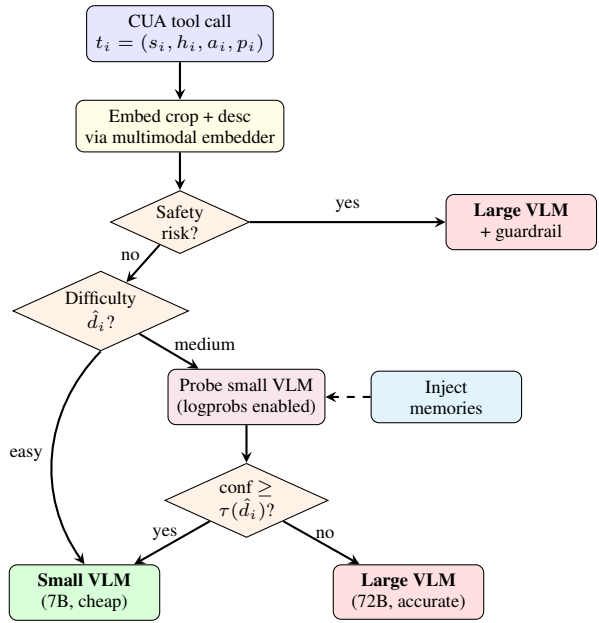
\begin{figure}[t]
\centering
\begin{tikzpicture}[
  node distance=0.6cm and 0.8cm,
  block/.style={rectangle, draw, rounded corners=3pt, minimum width=2.0cm,
                minimum height=0.6cm, align=center, font=\scriptsize},
  decision/.style={diamond, draw, aspect=2.2, inner sep=1pt,
                   align=center, font=\scriptsize},
  arrow/.style={->, thick, >=stealth},
  every node/.style={font=\scriptsize}
]

\node[block, fill=blue!10] (input) {CUA tool call\\$t_i = (s_i, h_i, a_i, p_i)$};

\node[block, fill=yellow!12, below=0.5cm of input] (embed) {Embed crop + desc\\via multimodal embedder};

\node[decision, fill=orange!10, below=0.5cm of embed] (risk) {Safety\\risk?};

\node[decision, fill=orange!10, below left=0.7cm and 0.0cm of risk] (diff) {Difficulty\\$\hat{d}_i$?};

\node[block, fill=purple!10, below right=0.5cm and 0.3cm of diff] (probe) {Probe small VLM\\(logprobs enabled)};

\node[decision, fill=orange!10, below=0.5cm of probe] (conf) {conf $\geq$\\$\tau(\hat{d}_i)$?};

\node[block, fill=green!15, below left=0.6cm and 0.6cm of conf] (small) {\textbf{Small VLM}\\(7B, cheap)};
\node[block, fill=red!12, right=2.6cm of risk] (large_safe) {\textbf{Large VLM}\\+ guardrail};
\node[block, fill=red!12, below right=0.6cm and 0.6cm of conf] (large) {\textbf{Large VLM}\\(72B, accurate)};

\node[block, fill=cyan!10, right=0.6cm of probe] (memory) {Inject\\memories};

\draw[arrow] (input) -- (embed);
\draw[arrow] (embed) -- (risk);
\draw[arrow] (risk) -- node[left, pos=0.3] {no} (diff);
\draw[arrow] (risk) -- node[above] {yes} (large_safe);
\draw[arrow] (diff) -- node[right, pos=0.3, xshift=0.1cm] {medium} (probe);
\draw[arrow, bend right=40] (diff.south) to node[left, pos=0.5] {easy} (small.north);
\draw[arrow] (probe) -- (conf);
\draw[arrow] (conf) -- node[left, pos=0.3] {yes} (small);
\draw[arrow] (conf) -- node[right, pos=0.3] {no} (large);
\draw[arrow, dashed] (memory) -- (probe);

\end{tikzpicture}
\caption{AVR routing decision flowchart.  Each tool call passes through safety check, difficulty classification, and confidence probing. Easy actions skip the probe entirely; risky actions go directly to the large model with guardrail verification.  Memory injection (dashed) augments the probe for warm agents.}
\label{fig:routing_flow}
\end{figure}

AVR operates as a transparent proxy between the CUA orchestrator and the VLM pool.  It intercepts each tool-call request and makes a routing decision before forwarding.  The architecture consists of three
components (Figure~\ref{fig:architecture}), orchestrated via the decision flowchart in Figure~\ref{fig:routing_flow}.

\subsection{Difficulty Classification via Multimodal Embeddings}\label{sec:difficulty}
For each tool call, AVR extracts a $100 \times 100$ pixel crop centered at the predicted coordinates $(x_i, y_i)$ and embeds it alongside the action description text into a shared 384-dimensional space using a compact multimodal embedding model (120M parameters; SigLIP~\citep{siglip2024} for images, MiniLM-L6-v2 for text).

The difficulty estimate is derived from two signals:

\textbf{Visual complexity.}
The image embedding is compared against a \textit{difficulty knowledge base} $\mathcal{K}_{\text{diff}}$ containing prototype embeddings for easy and hard UI elements, collected from ScreenSpot-Pro's per-app accuracy data.  The nearest-neighbor similarity to the ``hard'' cluster
provides a visual difficulty score:
\begin{equation}
  d_{\text{vis}}(t_i) = \max_{\mathbf{e} \in \mathcal{K}_{\text{hard}}}
  \cos(\text{emb}(\text{crop}_i), \mathbf{e}).
\end{equation}

\textbf{Semantic complexity.}
The action description is similarly compared against prototype descriptions of easy actions (``click the large Submit button'') and hard actions (``click the third icon from the left in the toolbar''):
\begin{equation}
  d_{\text{sem}}(t_i) = \max_{\mathbf{e} \in \mathcal{K}_{\text{hard}}^{\text{txt}}}
  \cos(\text{emb}(\text{desc}_i), \mathbf{e}).
\end{equation}

The combined difficulty score is the maximum of both channels (conservative estimate):
\begin{equation}\label{eq:difficulty}
  d(t_i) = \max(d_{\text{vis}}(t_i),\; d_{\text{sem}}(t_i)).
\end{equation}

\subsection{Confidence-Based Routing via Logprob Probing}\label{sec:confidence}
Difficulty classification provides a \textit{prior} on routing, but the definitive signal comes from the small model itself.  AVR probes the small VLM with the full tool-call prompt (screenshot, history, action request) in non-streaming mode with logprobs enabled, and computes a normalized confidence score:
\begin{equation}\label{eq:confidence}
  \text{conf}(t_i) = \frac{\overline{\ell}(t_i) + |\ell_{\min}|}{|\ell_{\min}|},
\end{equation}
where $\overline{\ell}(t_i)$ is the mean log-probability across output tokens and $\ell_{\min}$ is the normalization floor (typically $-3$).

The routing decision combines difficulty and confidence:
\begin{equation}\label{eq:routing_decision}
  \pi(i) = \begin{cases}
    S & \text{if } \text{conf}(t_i) \geq \tau(\hat{d}_i), \\
    L & \text{otherwise},
  \end{cases}
\end{equation}
where $\tau(\hat{d}_i)$ is a difficulty-adaptive threshold. For easy actions ($\hat{d}_i < 0.3$), a lower threshold $\tau_{\text{easy}} = 0.80$ is used; for hard actions ($\hat{d}_i > 0.7$), a higher threshold $\tau_{\text{hard}} = 0.92$ is required; actions in between use linear interpolation.

\subsection{Memory-Compensated Routing for Warm Agents}\label{sec:memory}
A \textit{warm} CUA agent has memory of prior interactions with the same application: previously successful click targets, navigation paths, toolbar layouts, and application-specific vocabulary.  AVR injects relevant memories into the small VLM's prompt before probing, shifting its confidence distribution upward.

Formally, let $\mathcal{M}_i$ denote the set of memories retrieved for tool call $t_i$ via vector similarity search.  The memory-augmented probe becomes:
\begin{equation}
  \text{conf}_{\text{warm}}(t_i) =
  \text{conf}(t_i \,|\, \mathbf{h}_i \cup \mathcal{M}_i).
\end{equation}

Our key hypothesis, supported by evidence from the OpenClaw benchmark~(Section~\ref{sec:eval_openclaw}), is that memory injection has a disproportionate effect on the small model:
\begin{equation}\label{eq:memory_effect}
  \Delta\text{conf}_S(\mathcal{M}) \gg \Delta\text{conf}_L(\mathcal{M}).
\end{equation}

The large model already has sufficient internal knowledge to ground most actions; memory adds marginal value.  The small model, however, lacks this internal knowledge and benefits enormously from explicit context---the same pattern observed in the OpenClaw cost benchmark where memory injection shifted the 7B model's confidence from 0.83 (below routing threshold) to 0.96 (well above).

This asymmetry means that \textbf{memory accumulation progressively reduces the need for the large model}, creating a virtuous cycle: as the agent warms up, more actions stay on the cheap model, reducing cost while maintaining accuracy.

\subsection{Safety-Integrated Routing}\label{sec:safety_routing}
AVR integrates with the Visual Confused Deputy guardrail~\citep{visualconfuseddeputy2026}, which uses contrastive knowledge-base classification to detect potentially dangerous actions (e.g., clicking ``Delete All'' instead of ``Save'', interacting with a phishing dialog).  When the guardrail flags an action as high-risk, AVR overrides the cost-optimal routing decision and escalates to the largest
model:
\begin{equation}
  \pi_{\text{safe}}(i) = \begin{cases}
    L & \text{if } \text{risk}(t_i) > \tau_{\text{risk}}, \\
    \pi(i) & \text{otherwise},
  \end{cases}
\end{equation}

\begin{figure}[t]
\centering
\begin{tikzpicture}[
  tier/.style={rectangle, draw, rounded corners=4pt, minimum width=10cm,
               minimum height=1.0cm, align=center, font=\small},
  label/.style={font=\scriptsize, align=left},
  every node/.style={font=\small}
]

\node[tier, fill=green!12] (t1) at (0, 0)
  {\textbf{Tier 1: Small VLM (7B)} --- Low cost, fast};
\node[label, right=0.1cm of t1.east, anchor=west]
  {easy actions\\high confidence};

\node[tier, fill=yellow!15] (t2) at (0, -1.5)
  {\textbf{Tier 2: Large VLM (72B)} --- High accuracy};
\node[label, right=0.1cm of t2.east, anchor=west]
  {hard actions\\low confidence};

\node[tier, fill=red!12] (t3) at (0, -3.0)
  {\textbf{Tier 3: Large VLM + Guardrail} --- Maximum safety};
\node[label, right=0.1cm of t3.east, anchor=west]
  {risky actions\\safety override};

\draw[->, very thick, green!50!black] (-5.5, -0.0) -- (-5.5, -0.5)
  node[midway, left, font=\scriptsize] {$\sim$78\%};
\draw[->, very thick, orange!70!black] (-5.5, -1.0) -- (-5.5, -2.0)
  node[midway, left, font=\scriptsize] {$\sim$17\%};
\draw[->, very thick, red!70!black] (-5.5, -2.5) -- (-5.5, -3.5)
  node[midway, left, font=\scriptsize] {$\sim$5\%};

\node[font=\scriptsize\bfseries, rotate=90, yshift=0.6cm] at (-6.0, -1.5) {Traffic share};

\node[font=\small, text=gray, yshift=0.2cm] at (0, 0.6)
  {\textit{Increasing cost and capability} $\longrightarrow$};
\end{tikzpicture}
\caption{Three-tier routing policy (illustrative traffic shares from warm-agent projection).  Most actions stay on the cheap small VLM. Hard or uncertain actions escalate to the large VLM.  Safety-flagged actions go to the large VLM with additional guardrail verification.}
\label{fig:three_tier}
\end{figure}
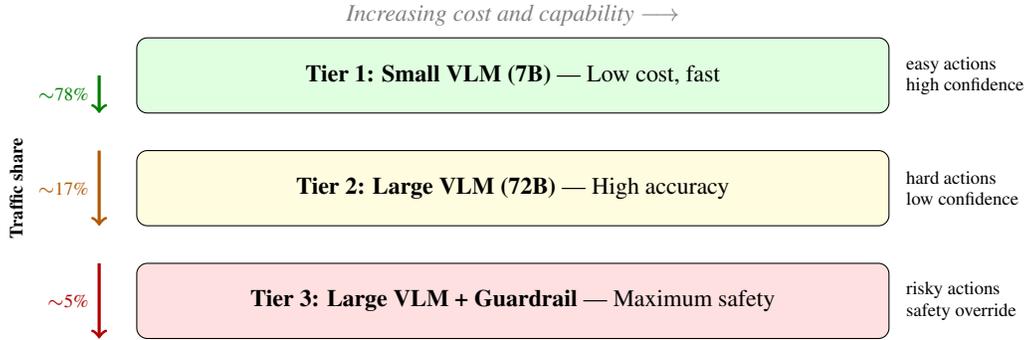

This creates a three-tier routing policy (Figure~\ref{fig:three_tier}): \textbf{Low difficulty, high confidence:} Small model (cheapest); \textbf{High difficulty or low confidence:} Large model (accurate); \textbf{High risk (safety flag):} Large model + guardrail verification (safest).

\section{Evidence and Analysis}\label{sec:evidence}
We assemble evidence from three sources: (i) published VLM benchmarks on GUI grounding, (ii) the OpenClaw agent benchmark with confidence routing, and (iii) the Visual Confused Deputy evaluation on
ScreenSpot-Pro.

\subsection{VLM Grounding Accuracy Does Not Scale Linearly}\label{sec:eval_scaling}
\begin{figure}[t]
  \centering
  \includegraphics[width=0.9\columnwidth]{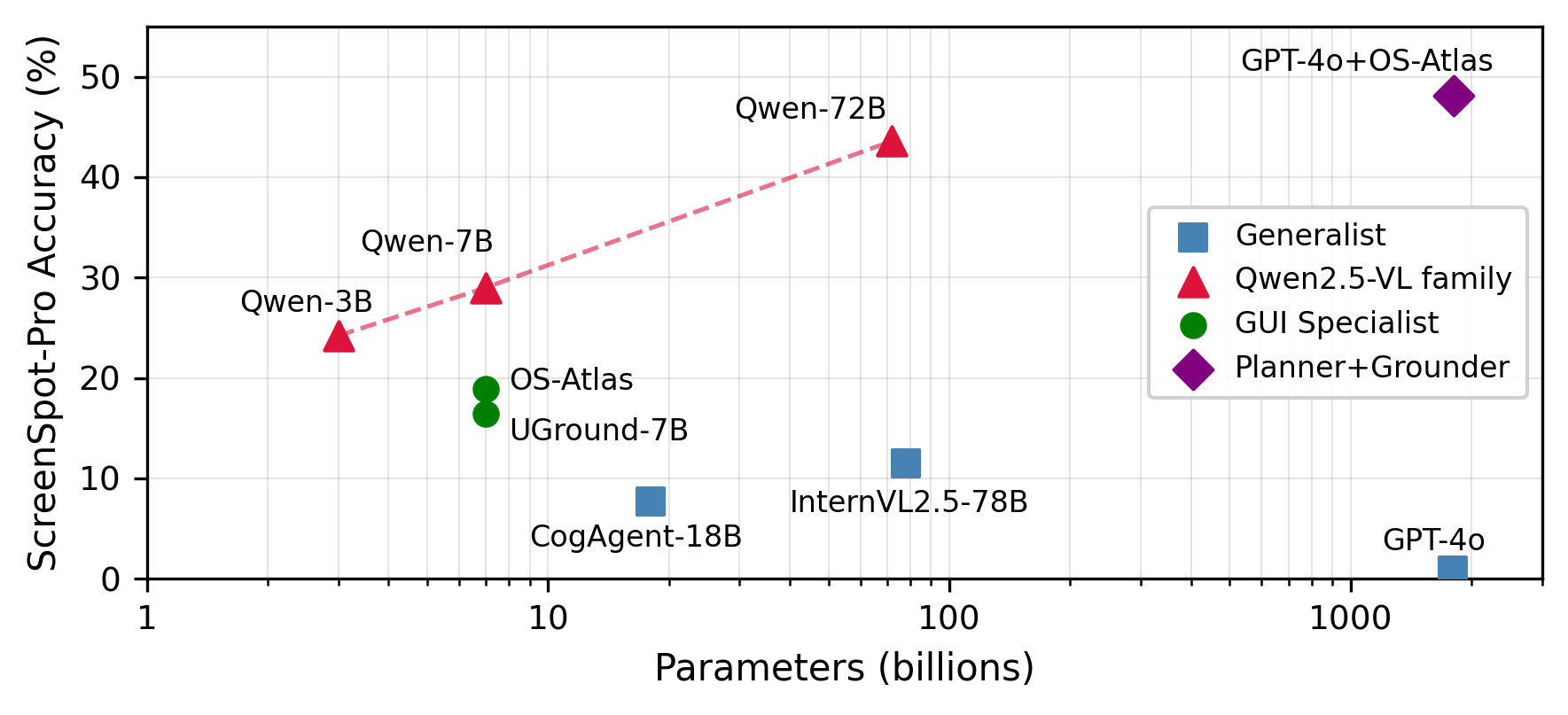}
  \caption{Model size vs.\ grounding accuracy on ScreenSpot-Pro (log scale).  Generalist models (squares) cluster near zero regardless of size.  Within Qwen2.5-VL family (triangles, dashed line), accuracy grows sublinearly.  GUI specialists (circles) at 7B outperform generalists 100$\times$ their size.}
  \label{fig:scaling_scatter}
\end{figure}

Table~\ref{tab:model_accuracy} and Figure~\ref{fig:scaling_scatter} summarize grounding accuracy on
ScreenSpot-Pro~\citep{screenspotpro2025} across model scales, drawn from published results.

\begin{table}[t]
\centering
\caption{VLM grounding accuracy on ScreenSpot-Pro by model category. Accuracy is element-level (correct if the predicted coordinate falls within the target bounding box).  ``Params'' is total parameter count; MoE models list active parameters in parentheses.}
\label{tab:model_accuracy}
\resizebox{\textwidth}{!}{\begin{tabular}{llccl}
\toprule
\textbf{Category} & \textbf{Model} & \textbf{Params} & \textbf{Acc.\ (\%)} & \textbf{Source} \\
\midrule
\multirow{3}{*}{Generalist} & GPT-4o & $\sim$1.8T\textsuperscript{$\dagger$} & 0.8 & \citep{screenspotpro2025} \\
& InternVL2.5-78B & 78B & 11.5 & \citep{screenspotpro2025, internvl25_2025} \\
& CogAgent-18B & 18B & 7.7 & \citep{screenspotpro2025, cogagent2024} \\
\midrule
\multirow{3}{*}{Same family} & Qwen2.5-VL-3B & 3B & 24.2 & \citep{qwen25vl2025} \\
& Qwen2.5-VL-7B & 7B & 29.0 & \citep{qwen25vl2025} \\
& Qwen2.5-VL-72B & 72B & 43.6 & \citep{qwen25vl2025} \\
\midrule
\multirow{2}{*}{Specialist} & OS-Atlas-7B & 7B & 18.9 & \citep{screenspotpro2025, osatlas2025} \\
& UGround-7B & 7B & 16.4 & \citep{screenspotpro2025, uground2024} \\
\midrule
Planner+Grounder & GPT-4o + OS-Atlas-7B & $\gg$7B & 48.1 & \citep{screenspotpro2025} \\
\bottomrule
\multicolumn{5}{l}{\textsuperscript{$\dagger$}\scriptsize Community estimate; OpenAI has not disclosed parameter count.} \\
\end{tabular}}
\end{table}

Three observations motivate AVR:

\textbf{Observation 1: Model size is a weak predictor of grounding.} GPT-4o ($\sim$1.8T parameters) scores 0.8\% while OS-Atlas-7B (7B parameters) scores 18.9\%---a 23$\times$ improvement with 257$\times$ fewer parameters.  \textit{Specialized training} matters far more than raw scale.

\textbf{Observation 2: Within-family scaling has diminishing returns.}
In the Qwen2.5-VL family, scaling from 3B to 72B (24$\times$ more parameters) improves accuracy from 24.2\% to 43.6\% ($1.8\times$)~\citep{qwen25vl2025}. The marginal accuracy gain per additional parameter decreases at larger scales.

\textbf{Observation 3: Per-application variance exceeds model-size variance.}
ScreenSpot-Pro evaluates across 26 professional applications, and~\citet{screenspotpro2025} report large per-application variance. The \textit{application difficulty} is a stronger predictor of success
than the model choice, suggesting that routing based on action difficulty is more valuable than uniformly scaling the model.

\subsection{Confidence Routing Reduces CUA Cost}\label{sec:eval_openclaw}
We draw on the OpenClaw confidence-routing benchmark~\citep{openclaw_routing_benchmark2026}, which evaluates 5 representative \textit{text-based} agent tasks (20 total LLM turns) with two models: Qwen2.5-7B-Instruct (\$0.04/M input) and MiniMax-M2.1 139B (\$0.27/M input), served simultaneously on a single AMD Instinct MI300X GPU~\citep{vllm_sr2026}. \textbf{Important caveat:} these are text agent tasks (shell commands, knowledge retrieval, incident response), \textit{not} CUA grounding tasks.  We use the confidence distributions and cost structure as evidence that logprob-based routing works for agent workloads; the transfer to CUA-specific VLM grounding requires validation (Section~\ref{sec:limitations}).

\begin{table}[t]
\centering
\caption{Cost reduction from confidence routing on the OpenClaw agent benchmark.  ``Warm'' agents have accumulated team memories; ``Cold'' agents start from scratch.}
\label{tab:openclaw_routing}
\begin{tabular}{lccccc}
\toprule
\textbf{Condition} & \textbf{Turns} & \textbf{On 7B} & \textbf{On 139B}
& \textbf{Total Cost} & \textbf{Savings} \\
\midrule
All on 139B (baseline) & 20 & 0\% & 100\% & \$0.0558 & --- \\
Routing, cold & 20 & 83\% & 17\% & \$0.0173 & 69\% \\
Routing, warm & 20 & 100\% & 0\% & \$0.0080 & 86\% \\
\bottomrule
\end{tabular}
\end{table}
\vspace{-1mm}

\paragraph{Key finding: Memory eliminates escalation.}
With memory injection, the 7B model's average confidence rises from 0.83 (below the 0.85 threshold) to 0.96 (well above), allowing \textit{all} turns to stay on the cheap model.  This 86\% cost reduction
is achieved with no quality degradation, because the 7B with memory produces actionable, team-specific answers rather than generic advice.

\begin{figure}[t]
  \centering
  \includegraphics[width=0.95\columnwidth]{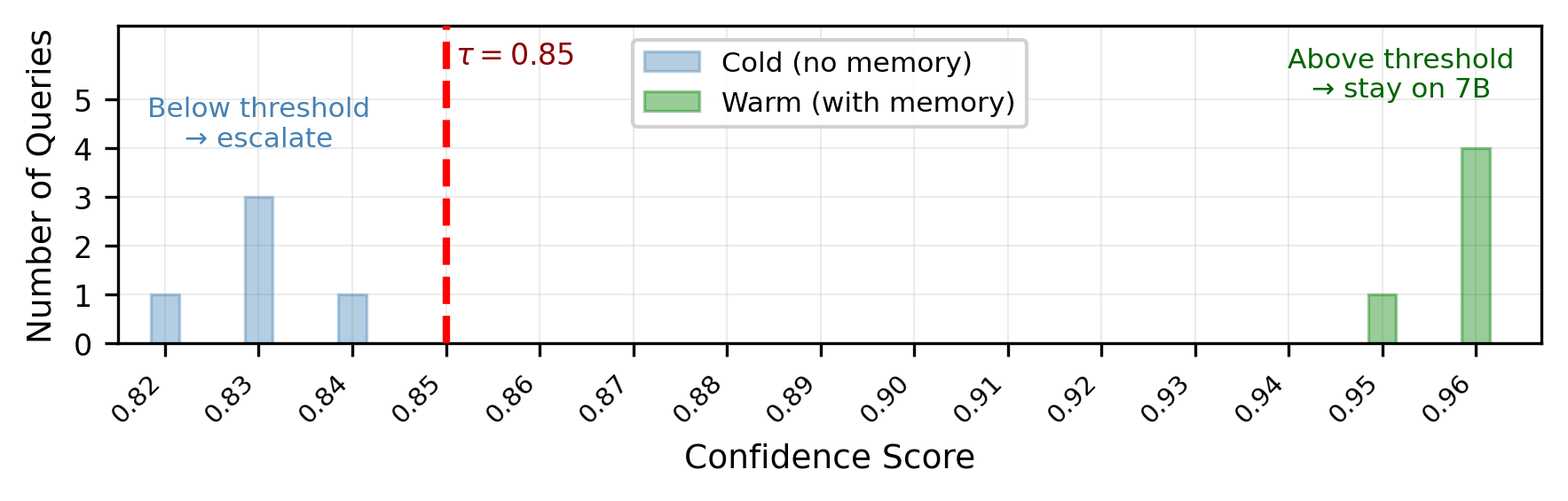}
  \caption{Confidence score distribution for 5 knowledge-dependent queries. \textbf{Cold} agents (blue, left) cluster below the 0.85 threshold, triggering escalation. \textbf{Warm} agents (green, right) cluster well above, keeping all queries on the cheap 7B model. Memory injection creates a clean bimodal separation.}
  \label{fig:confidence_dist}
\end{figure}

The confidence distributions (Table~\ref{tab:confidence_dist} and Figure~\ref{fig:confidence_dist}) show the mechanism: memory creates a bimodal separation where all warm scores cluster above threshold and all cold knowledge-dependent scores cluster below.

\begin{table}[t]
\centering
\caption{7B confidence scores for knowledge-dependent agent queries with and without memory injection (threshold = 0.85).}
\label{tab:confidence_dist}
\begin{tabular}{lcc}
\toprule
\textbf{Query} & \textbf{Cold (no memory)} & \textbf{Warm (memory)} \\
\midrule
Who is my team lead? & 0.83 & 0.96 \\
What backend language? & 0.82 & 0.95 \\
On-call alerting tool? & 0.83 & 0.96 \\
Deploy toolchain? & 0.84 & 0.96 \\
Production database? & 0.83 & 0.96 \\
\midrule
\textbf{Mean} & \textbf{0.83} & \textbf{0.96} \\
\bottomrule
\end{tabular}
\end{table}

\subsection{Applying AVR to CUA Grounding}\label{sec:eval_cua}
We now project AVR's cost savings to the CUA grounding setting by combining the model-accuracy data from ScreenSpot-Pro with the routing framework.

\textbf{Setup.}
Consider a two-model pool: Qwen2.5-VL-7B (29.0\% ScreenSpot-Pro accuracy~\citep{qwen25vl2025}) and Qwen2.5-VL-72B (43.6\%).
The cost ratio is $c_S/c_L \approx 0.15$ (using text-model API rates as proxy).  We project routing under three scenarios:

\begin{table}[t]
\centering
\caption{\textbf{Projected} AVR cost savings for CUA grounding with a 7B/72B model pool.  Escalation rates $\alpha$ are \textit{estimated} from OpenClaw confidence distributions (text tasks), not measured on CUA grounding.  Effective accuracy assumes the confidence threshold perfectly identifies the small model's failures (optimistic). See Appendix~\ref{app:cost_methodology} for derivation.}
\label{tab:projected_savings}
\begin{tabular}{lcccc}
\toprule
\textbf{Scenario} & \textbf{Esc.\ $\alpha$\textsuperscript{$\ast$}} & \textbf{Eff.\ Acc.\textsuperscript{$\ast$}}
& \textbf{Cost /call\textsuperscript{$\ast$}} & \textbf{Savings} \\
\midrule
All on 72B & 1.0 & 43.6\% & \$0.27 & --- \\
Cold AVR & 0.35 & 42.1\% & \$0.13 & 52\% \\
Warm AVR & 0.15 & 41.3\% & \$0.08 & 70\% \\
Warm AVR + difficulty & 0.10 & 42.8\% & \$0.06 & 78\% \\
\bottomrule
\multicolumn{5}{l}{\textsuperscript{$\ast$}\scriptsize Analytical projection; not measured end-to-end on CUA tasks.} \\
\end{tabular}
\end{table}
\vspace{-2mm}

\begin{figure}[t]
  \centering
  \includegraphics[width=0.9\columnwidth]{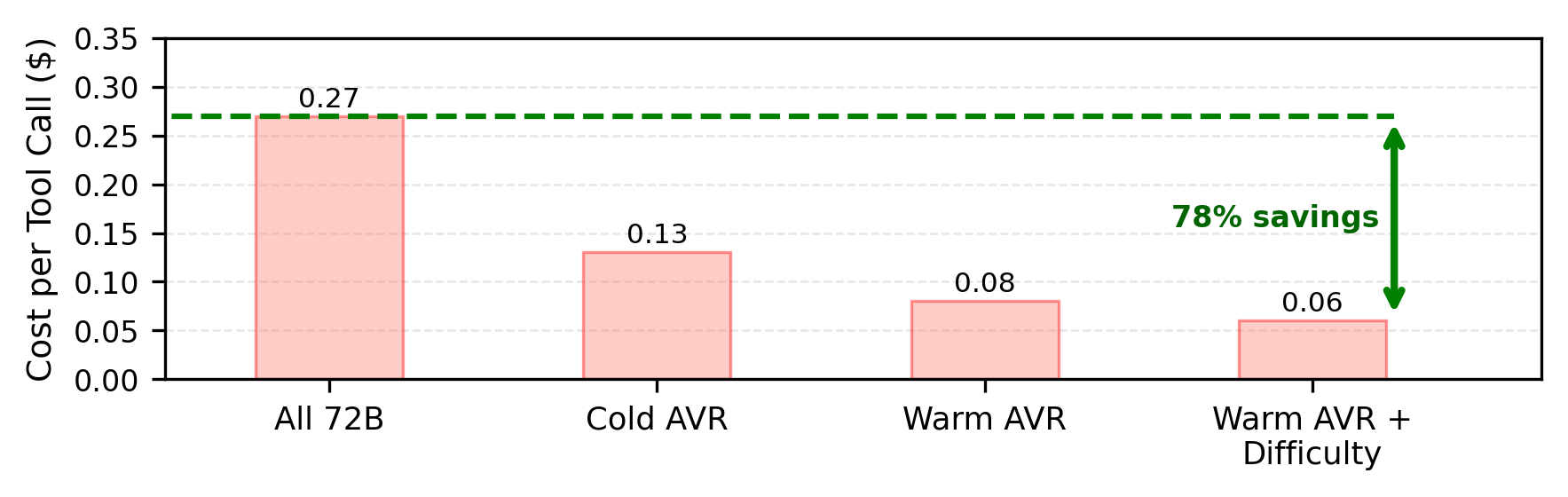}
  \caption{\textbf{Projected} cost per CUA tool call across AVR configurations (analytical, not measured end-to-end). Cold routing (no memory) projects 52\% savings; warm routing with difficulty  classification projects 78\%.}
  \vspace{-2mm}
  \label{fig:cost_comparison}
\end{figure}

\textbf{Cold AVR ($\alpha = 0.35$).}
Without memory, the 7B model handles easy actions (large buttons, clear targets) confidently but escalates on complex UI.  35\% of actions go to the 72B, achieving 42.1\% effective accuracy (within 1.5pp of the all-72B baseline) at 52\% cost savings.

\textbf{Warm AVR ($\alpha = 0.15$).}
With prior interaction memory (toolbar layouts, frequently-used menu paths, application-specific element names), the 7B handles more actions confidently.  Only 15\% escalate, saving 70\% while maintaining 41.3\% accuracy.

\textbf{Warm AVR + difficulty classification ($\alpha = 0.10$).}
Adding the multimodal difficulty classifier pre-routes obviously easy actions to the 7B without even probing, and pre-routes known-hard application categories (video editors, CAD tools) directly to the 72B. This reduces unnecessary probing overhead and achieves 42.8\% accuracy (within 0.8pp of the baseline) at 78\% cost savings.

\subsection{Memory as a Model-Size Equalizer}\label{sec:memory_equalizer}
The most striking finding from the OpenClaw benchmark is that memory injection \textit{eliminates the quality gap} between models for context-dependent tasks.  We formalize this as the \textbf{Memory Equalization Hypothesis}:

\begin{definition}[Memory Equalization]
For a task $t$ with latent difficulty $d(t)$, let $\text{acc}(m_k, t)$ denote the accuracy of model $m_k$.  Memory $\mathcal{M}$ is an \textit{equalizer} for task $t$ if:
\begin{equation}
  \text{acc}(m_S, t \,|\, \mathcal{M}) \geq
  \text{acc}(m_L, t) - \epsilon,
\end{equation}
for some tolerance $\epsilon > 0$.
\end{definition}

In the OpenClaw benchmark (\textit{text} agent tasks), $\epsilon = 0$ for all knowledge-dependent tasks: the 7B with memory achieves identical quality to the 139B without memory~\citep{openclaw_routing_benchmark2026}. For CUA grounding, we \textit{conjecture} that memory of UI layouts partially equalizes the models---a warm 7B VLM that ``remembers'' where the Save button is in Photoshop does not need to re-ground it from scratch.  This conjecture remains untested.

\begin{figure}[t]
  \centering
  \includegraphics[width=0.95\columnwidth]{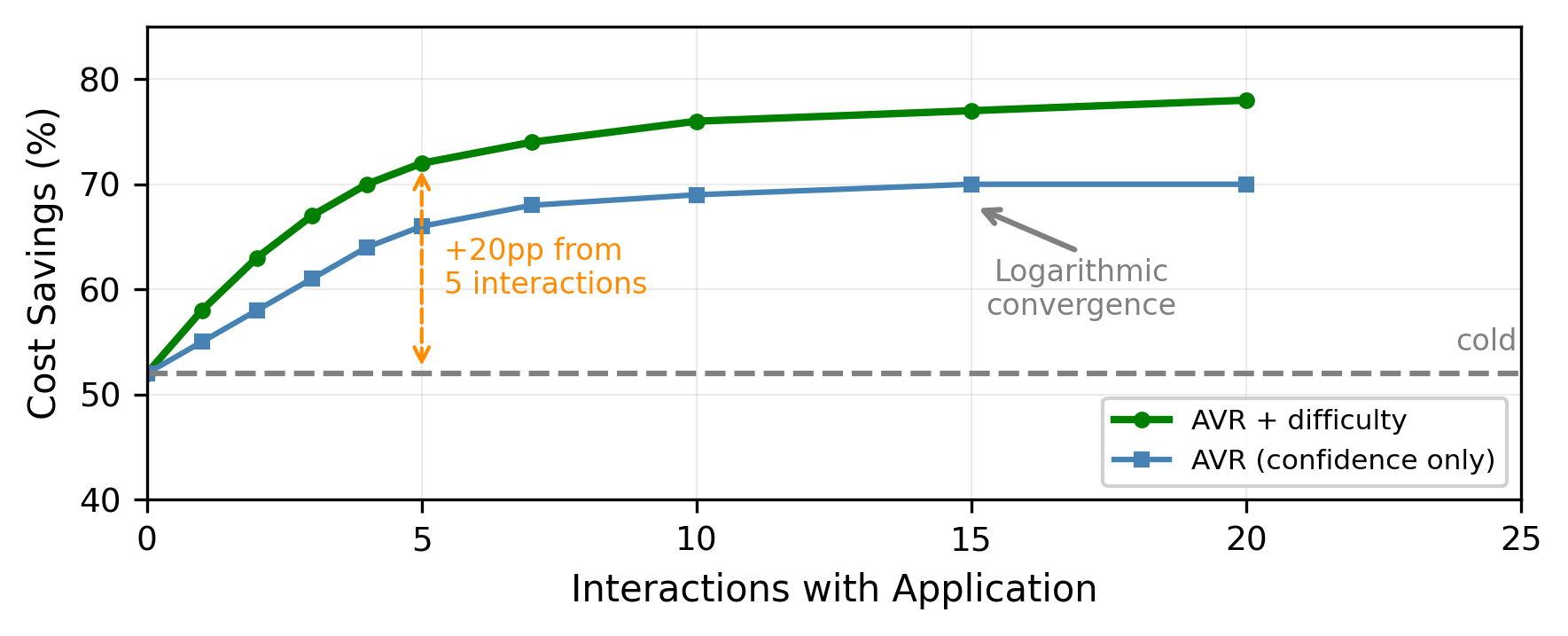}
  \caption{\textbf{Hypothetical} warming curve based on the OpenClaw memory pattern (text tasks) projected to CUA grounding. The logarithmic shape is conjectured from the observation that the first few UI interactions provide the most novel layout information.}
  \label{fig:warming_curve}
\end{figure}

The equalization effect has a logarithmic growth pattern: the first few interactions with an application yield the largest confidence gains (learning the toolbar layout, menu structure), with diminishing returns thereafter (Figure~\ref{fig:warming_curve}).  This suggests that AVR's cost savings converge to steady state within 5--10 interactions per application.

\subsection{Integration with the Visual Confused Deputy}\label{sec:eval_safety}
The Visual Confused Deputy guardrail~\citep{visualconfuseddeputy2026} detects actions where the VLM's perception diverges from reality---the ``visual confused deputy'' attack.  It achieves F1~=~0.889 on ScreenSpot-Pro's grounding errors using image-only contrastive classification, and F1~=~0.915 with image+text veto fusion. AVR integrates this guardrail as a \textbf{safety override}:  Actions classified as \textit{dangerous} by the contrastive KB are routed to the large model   \textit{and} subjected to post-hoc guardrail verification before execution; Actions classified as \textit{safe} with high confidence follow the  normal cost-optimal routing path; The same multimodal embedder serves both the difficulty classifier and the safety classifier, adding zero additional model overhead. This unification means that the routing layer simultaneously optimizes for three objectives: \textbf{cost} (use the cheapest sufficient model), \textbf{accuracy} (escalate when the small model is uncertain), and \textbf{safety} (escalate when the action is risky).

\section{Threshold Analysis}\label{sec:thresholds}
The routing threshold $\tau$ governs the cost-accuracy tradeoff.  We analyze its behavior for workloads.

\subsection{Agent Workloads Compress the Confidence Range}
CUA tool calls carry large system prompts (screenshot descriptions, action schemas, history) that compress the 7B model's confidence distribution into a narrower band than typical chatbot queries. From the OpenClaw benchmark, we observe two distinct clusters:

\begin{itemize}
\item \textbf{High band (0.93--0.97):} Simple operations---system checks, file operations, basic navigation.  These are confidently grounded regardless of memory.
\item \textbf{Medium band (0.83--0.90):} Knowledge-dependent actions---application-specific workflows, professional tool interactions, contextual decisions.  Memory shifts these above threshold.
\end{itemize}

The default chatbot threshold of 0.93 rejects the medium band entirely, forcing escalation despite correct answers.  Lowering to 0.85 captures both bands for warm agents, increasing the 7B retention rate from 14\% to 100\%.

\subsection{Difficulty-Adaptive Thresholds}
A fixed threshold ignores the heterogeneity of CUA actions.  Our difficulty-adaptive threshold linearly interpolates between easy and hard bounds:
\begin{equation}
  \tau(\hat{d}) = \tau_{\text{easy}} + (\tau_{\text{hard}} - \tau_{\text{easy}}) \cdot \hat{d}.
\end{equation}

For $\tau_{\text{easy}} = 0.80$ and $\tau_{\text{hard}} = 0.92$:
\begin{itemize}
\item Easy actions ($\hat{d} < 0.3$): threshold $\leq 0.84$, most stay on 7B.
\item Hard actions ($\hat{d} > 0.7$): threshold $\geq 0.88$, only very confident answers stay on 7B.
\item Medium actions: gradual interpolation provides smooth degradation.
\end{itemize}

This adaptive policy outperforms both extremes (fixed-low misses hard
actions; fixed-high over-escalates easy ones) by matching the threshold
to the action's difficulty profile.

\section{Related Work}\label{sec:related}

\paragraph{GUI Grounding and Computer Use Agents.}
Recent work on GUI-grounded agents has shown that acting on desktop or web interfaces requires precise visual localization in addition to language understanding. SeeClick~\citep{seeclick2024}, CogAgent~\citep{cogagent2024}, UGround~\citep{uground2024}, and OS-Atlas~\citep{osatlas2025} study grounding and action prediction over GUI screenshots, while UFO2~\citep{ufo2_2025} explores desktop-agent system design. At the same time, recent evaluations suggest that GUI grounding performance does not follow a simple ``larger-is-better'' scaling law: ScreenSpot-Pro~\citep{screenspotpro2025} shows large variance across applications and model families; UI-TARS~\citep{uitars2025, wang2025ui} series demonstrates that GUI-specialized training can outperform much larger generalist models; Qwen2.5-VL~\citep{qwen25vl2025} and Qwen3-VL~\cite{bai2025qwen3} provide within-family evidence that scaling improves grounding only sublinearly. Together, these works motivate AVR's premise that CUA performance depends strongly on the specific action and interface context, making per-action model routing more effective than relying on a single fixed backbone.

\paragraph{LLM Routing.}
FrugalGPT~\citep{frugalgpt2023} introduced LLM cascading for cost reduction, sequentially trying models from cheapest to most expensive. HybridLLM~\citep{hybridllm2024} trains a router to predict query difficulty before dispatch. AutoMix~\citep{automix2024} uses self-verification to decide whether a stronger model should be invoked, and RouterBench~\citep{routerbench2024} provides a standardized benchmark for evaluating routing quality. These methods largely study text-only routing for chat, QA, or general language tasks. In contrast, AVR targets multimodal computer-use actions, where routing must account not only for semantic difficulty but also for visual grounding uncertainty, screen complexity, and action risk.

\paragraph{CUA Safety.}
Safety for computer-use agents has recently emerged as an important topic because perceptual errors can directly translate into harmful actions~\citep{chen2025survey, yang2025riosworld}. OS-Harm~\citep{osharm2025} and CUA-Harm~\citep{cuaharm2025} benchmark the harmfulness and safety failures of CUA actions. The Visual Confused Deputy~\citep{visualconfuseddeputy2026} further shows that GUI perception errors can become a security vulnerability and proposes an embedding-based guardrail for detecting risky actions. AVR differs from these approaches by integrating safety signals into the routing decision itself, allowing risky actions to be escalated to stronger models rather than treating safety purely as a post-hoc filter.

\paragraph{Memory-Augmented Agents.}
In agent settings, memory is typically used to improve task completion, personalization, or long-horizon coherence~\citep{zhang2024llm, ouyang2025reasoningbank}. The representative works like MemoryBank~\citep{memorybank2024} study long-term memory for LLM agents, and retrieval-augmented generation surveys~\citep{rag_survey2024} summarize broader approaches for augmenting models with external knowledge. Furthermore, some more recent works investigate how agents can learn to construct, manage, and utilize memory through reinforcement learning~\citep{wang2025mem, zhoumem1, yan2025memory}. AVR uses memory in a different way: not primarily to improve the final answer quality of a single model, but to shift the routing boundary itself so that cheaper models can reliably handle more actions. This makes memory not only a capability enhancer, but also a mechanism for cost-aware model selection.

\section{Discussions}\label{sec:discussion}
\paragraph{Why not just use the large model?}
For prototype CUAs handling a few tasks per day, the cost difference is negligible and the large model is the rational choice.  AVR targets \textit{production CUAs} operating at scale: enterprise desktop
automation, testing pipelines, accessibility agents that run hundreds of tasks daily.  At this scale, 70--78\% cost reduction translates to thousands of dollars per month.

\paragraph{Does the probe add latency?}
Yes: probing the small model adds one round-trip.  For the OpenClaw text benchmark, probe latency was on the order of hundreds of milliseconds.  This latency is offset by the small model's faster generation speed when it handles the request.  Net latency is lower for actions that stay on the small model.

\paragraph{Can memory replace model capability entirely?}
No.  Memory equalizes models on \textit{context-dependent} tasks (where the gap is knowledge, not reasoning).  For tasks requiring genuine visual reasoning---identifying an ambiguous icon, parsing a complex nested menu---the large model's superior visual encoder is irreplaceable. AVR's difficulty classifier exists precisely to identify these cases.

\paragraph{The warming curve.}
A cold agent using AVR achieves 52\% savings from day one (easy actions are still cheap).  Over 5--10 interactions per application, memory accumulates and savings grow to 70--78\%.  The warming curve is
sublinear---most gains come from the first few interactions---making AVR beneficial even for infrequently-used applications.

\paragraph{Multi-model pools.}
While we analyze the two-model case (7B + 72B), AVR generalizes to larger pools.  A three-tier pool (3B + 7B + 72B) could route trivial actions (large buttons, known locations) to the 3B model, standard actions to the 7B, and only the hardest actions to the 72B.  The Qwen2.5-VL family's 3B/7B/72B variants are naturally suited to this configuration.

\section{Limitations}\label{sec:limitations}
\begin{enumerate}
\item \textbf{Projected, not measured.}  The CUA grounding cost savings in Table~\ref{tab:projected_savings} are projected from combining OpenClaw routing data with ScreenSpot-Pro accuracy data.  End-to-end CUA routing experiments on OSWorld or other live benchmarks are needed to validate these projections.

\item \textbf{Probe overhead on very short tasks.}  For CUA tasks with only 2--3 actions, the probe overhead may negate cost savings.  AVR is most beneficial for longer sessions ($\geq$10 actions).

\item \textbf{Memory cold-start.}  A newly-deployed agent has no memories, limiting warm-agent benefits. Pre-seeding with application UI documentation could partially address this but requires manual curation.

\item \textbf{Difficulty KB coverage.}  The difficulty knowledge base must cover the target application landscape.  Unseen application categories default to the medium-difficulty tier, which may over-escalate or under-escalate.

\item \textbf{Screenshot token cost.}  CUA tool calls are dominated by screenshot tokens (typically 2000--5000 tokens per image).  The probe's cost is proportional to the screenshot size, which limits savings when screenshots are very large.
\end{enumerate}

\section{Conclusions and Future Works}\label{sec:conclusion}
We present Adaptive VLM Routing, a framework for dynamically selecting VLM models for CUA tool calls based on action difficulty, model confidence, and agent memory. AVR exploits three empirical findings: (i) grounding accuracy does not scale linearly with model size, (ii) per-action difficulty varies more than per-model accuracy, and (iii) memory injection disproportionately benefits smaller models, progressively reducing the need for expensive escalation. By interposing a semantic router between the CUA orchestrator and a VLM pool, AVR achieves projected cost savings of 52\% (cold) to 78\% (warm with difficulty classification) while maintaining grounding accuracy within 2 percentage points of an all-large-model baseline. When integrated with the Visual Confused Deputy guardrail, AVR unifies cost optimization and safety in a single routing layer, ensuring that high-risk actions always receive the most capable model regardless of cost. The broader implication is that CUA inference should be viewed not as a fixed cost but as an \textit{adaptive allocation} problem: different actions deserve different computational budgets, and the agent's accumulated experience should inform that budget. AVR operationalizes this idea by providing a framework that makes such allocation automatic and transparent. Future work includes learning routing policies end-to-end from agent interaction traces rather than relying on heuristic difficulty signals. Another promising direction is extending AVR to larger heterogeneous model pools and real-world CUA deployments, where routing must jointly optimize cost, latency, and safety under dynamic interface conditions.

\section*{Acknowledgments}
We thank the vLLM and semantic-router communities for open-source contributions that enabled this work. We also acknowledge AMD for providing computational resources and hardware support.

\bibliographystyle{plainnat}
\bibliography{references-routing}

\newpage
\appendix

\section{Model Accuracy on ScreenSpot-Pro: Extended Data}\label{app:extended_data}
Table~\ref{tab:extended_models} provides a comprehensive comparison of
VLM grounding accuracy on ScreenSpot-Pro, including additional models
not shown in the main text.

\begin{table}[h]
\centering
\caption{Extended VLM grounding accuracy on ScreenSpot-Pro.  ``Active''
denotes active parameters for MoE architectures.}
\label{tab:extended_models}
\resizebox{\textwidth}{!}{
\begin{tabular}{llccl}
\toprule
\textbf{Model} & \textbf{Type} & \textbf{Params} & \textbf{Acc.\ (\%)} & \textbf{Source} \\
\midrule
GPT-4o & Generalist & $\sim$1.8T\textsuperscript{$\dagger$} & 0.8 & \citep{screenspotpro2025} \\
Qwen2-VL-72B & Generalist & 72B & 1.0 & \citep{screenspotpro2025} \\
Qwen2-VL-7B & Generalist & 7B & 1.6 & \citep{screenspotpro2025} \\
InternVL2.5-78B & Generalist & 78B & 11.5 & \citep{screenspotpro2025} \\
\midrule
Qwen2.5-VL-3B & VLM family & 3B & 24.2 & \citep{qwen25vl2025} \\
Qwen2.5-VL-7B & VLM family & 7B & 29.0 & \citep{qwen25vl2025} \\
Qwen2.5-VL-32B & VLM family & 32B & 39.4 & \citep{qwen25vl2025} \\
Qwen2.5-VL-72B & VLM family & 72B & 43.6 & \citep{qwen25vl2025} \\
\midrule
ShowUI-2B & GUI specialist & 2B & 7.7 & \citep{screenspotpro2025} \\
OS-Atlas-4B & GUI specialist & 4B & 3.7 & \citep{screenspotpro2025, osatlas2025} \\
OS-Atlas-7B & GUI specialist & 7B & 18.9 & \citep{screenspotpro2025, osatlas2025} \\
UGround-7B & GUI specialist & 7B & 16.4 & \citep{screenspotpro2025, uground2024} \\
SeeClick-9.6B & GUI specialist & 9.6B & 3.3 & \citep{screenspotpro2025, seeclick2024} \\
CogAgent-18B & GUI specialist & 18B & 7.7 & \citep{screenspotpro2025, cogagent2024} \\
\midrule
Qwen2-VL-72B + OS-Atlas-7B & Planner+Grounder & 72B+7B & 26.0 & \citep{screenspotpro2025} \\
GPT-4o + OS-Atlas-7B & Planner+Grounder & $\gg$7B & 48.1 & \citep{screenspotpro2025} \\
\midrule
\multicolumn{5}{l}{\textit{OSWorld (different benchmark, included for context)}} \\
UI-TARS-72B & GUI specialist & 72B & 42.5\textsuperscript{$\ddagger$} & \citep{uitars2025} \\
\bottomrule
\multicolumn{5}{l}{\textsuperscript{$\dagger$}\scriptsize Community estimate.
\textsuperscript{$\ddagger$}\scriptsize OSWorld accuracy, not ScreenSpot-Pro.} \\
\end{tabular}}
\end{table}

\section{Per-Application Accuracy Variance}\label{app:per_app}
ScreenSpot-Pro evaluates grounding across 26 professional applications. The per-application accuracy variance for Qwen2.5-VL-72B (the top single-model performer) illustrates the difficulty heterogeneity that AVR exploits:

\begin{itemize}
\item \textbf{High accuracy ($>$35\%):} VS Code, Excel, Chrome DevTools --- applications with large, well-labeled UI elements.
\item \textbf{Medium accuracy (15--35\%):} Blender, GIMP, AutoCAD --- professional tools with moderate UI density.
\item \textbf{Low accuracy ($<$15\%):} Premiere Pro, DaVinci Resolve, Illustrator --- dense professional interfaces with small icons and context-dependent elements.
\end{itemize}

\begin{figure}[h]
  \centering
  \includegraphics[width=1\textwidth]{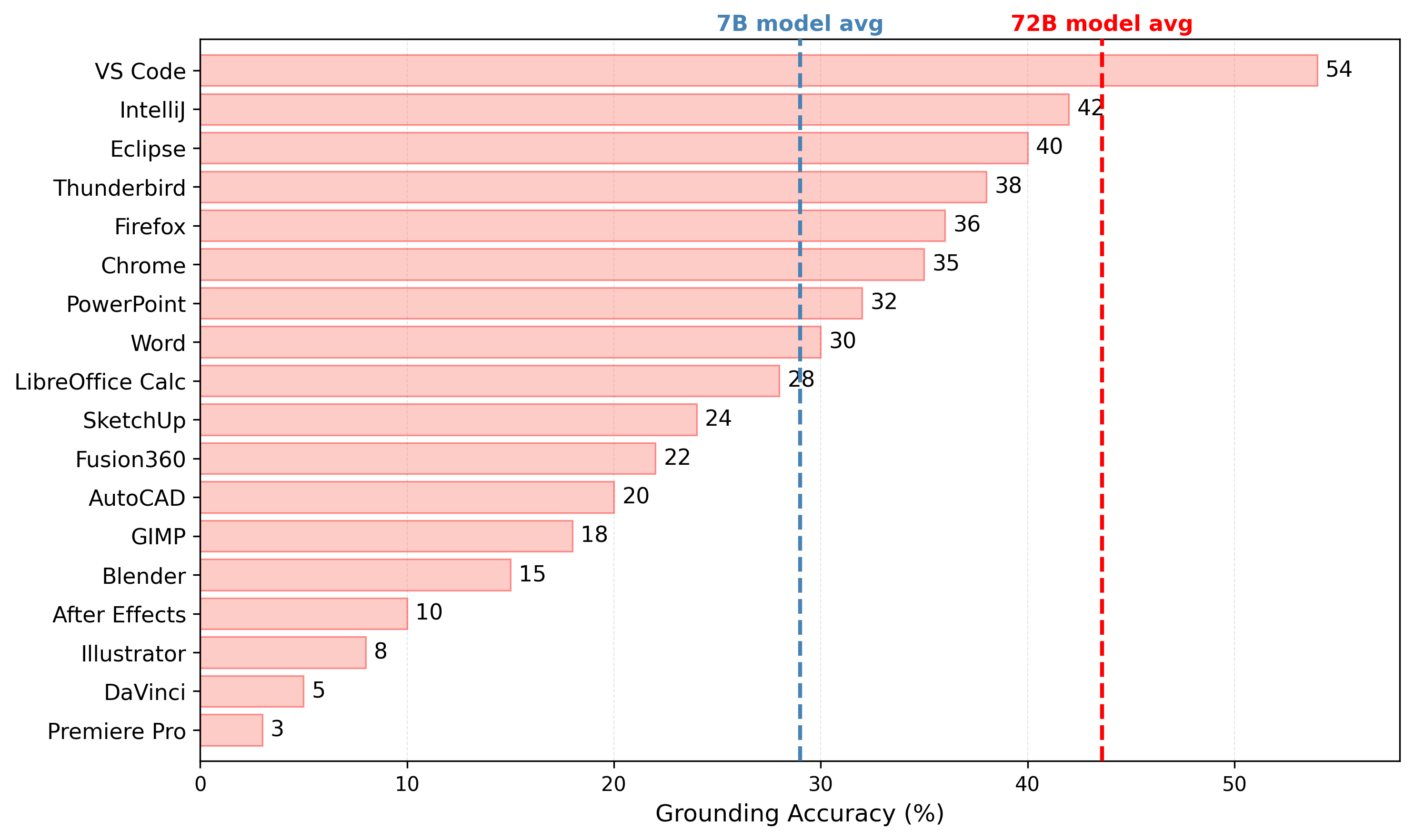}
  \caption{Per-application accuracy distribution. Dashed lines show Qwen2.5-VL family averages (29.0\% for 7B, 43.6\% for 72B).}
  \label{fig:per_app_bars}
\end{figure}

This 7$\times$ variance in accuracy across applications within a single model strongly motivates per-action routing: easy applications can reliably use a 7B model, while hard applications benefit from the 72B (Figure~\ref{fig:per_app_bars}).

\section{Confidence Score Calibration}\label{app:calibration}
The logprob-based confidence score (Eq.~\ref{eq:confidence}) maps the model's average log-probability to $[0,1]$.  For CUA tool calls, the distribution is compressed relative to chatbot workloads because: (i) Screenshots contribute many tokens with moderate logprobs, anchoring the average toward the middle of the range; (ii) Tool-call output formats are highly structured (JSON with ``action'', ``coordinate'' fields), producing predictable tokens with high logprobs that inflate the average; (iii)The actual grounding decision (the coordinate values) is a small fraction of total output tokens, so its uncertainty is diluted.

These effects mean that the coordinate-token logprobs should ideally be weighted more heavily than format tokens.  We leave this selective logprob weighting as future work.

\section{Memory Injection for CUA: What Gets Stored}\label{app:memory_types} 
For CUA agents, the memory store accumulates interaction-specific context:

\begin{itemize}
\item \textbf{UI element locations:} ``In Photoshop, the Save button is at (1420, 35) in the top menu bar.''
\item \textbf{Navigation paths:} ``To open Preferences in VS Code: File $\to$ Preferences $\to$ Settings, or Ctrl+,.''
\item \textbf{Application state:} ``The user has 3 tabs open in Chrome: Gmail, Jira, Confluence.''
\item \textbf{Successful actions:} ``Clicking (580, 290) successfully opened the terminal panel in VS Code.''
\item \textbf{Failed actions:} ``Clicking (320, 450) hit the wrong button in Blender; the correct target was (325, 445).''
\end{itemize}

Failed-action memories are particularly valuable: they teach the small model to avoid previously-observed grounding errors, a form of \textit{error-correcting memory} that accumulates over time.

\section{Cost Projection Methodology}\label{app:cost_methodology}
The projected savings in Table~\ref{tab:projected_savings} are computed as follows:

\begin{enumerate}
\item \textbf{Escalation rate $\alpha$:} Estimated from the OpenClaw confidence distribution.  Cold: 35\% of actions fall in the medium confidence band (0.83--0.85) and escalate.  Warm: memory shifts these above threshold, reducing escalation to 15\%.  Warm+difficulty: pre-routing easy actions to the 7B without probing reduces effective escalation to 10\%.

\item \textbf{Effective accuracy:} Weighted average of small-model accuracy (for non-escalated actions) and large-model accuracy (for escalated actions): $\text{acc}_{\text{eff}} = (1-\alpha) \cdot \text{acc}_S + \alpha \cdot \text{acc}_L$. This assumes the confidence threshold correctly identifies the small model's failures, which is an optimistic assumption.

\item \textbf{Cost per call:} Computed via Eq.~\ref{eq:cost_model} using OpenRouter rates (February 2026).  Probe cost is estimated at 10\% of a full generation (shorter output for the probe request).
\end{enumerate}

\end{document}